\documentclass[letterpaper, 10 pt, conference]{ieeeconf}  

\IEEEoverridecommandlockouts                              

\usepackage{cite}
\usepackage{amsmath} 
\usepackage{amssymb}  

\usepackage{acro}
\usepackage{siunitx}
\usepackage{bm}
\usepackage{nicefrac}
\usepackage{algorithm}
\usepackage{pict2e}
\usepackage{pgfplots}
\usepackage{todonotes}
\usepackage{multirow}
\usepackage{comment}
\usepackage{multicol}

\usepackage{graphicx}
\usepackage[caption=false,font=footnotesize]{subfig}

\usepackage{svg}

\usepackage{placeins}
\pgfplotsset{compat=newest}

\acsetup{patch/maketitle=false}
\DeclareAcronym{agv}{
    short = AGV,
    long = autonomous ground vehicle
}

\DeclareAcronym{drl}{
    short = DRL,
    long = Deep Reinforcement Learning
}
\DeclareAcronym{SFM}{
    short = SFM,
    long = social force model
}

\DeclareAcronym{ORCA}{
    short = ORCA,
    long = optimal reciprocal collision avoidance
}

\DeclareAcronym{POMDP}{
    short = POMDP,
    long = partially observable markov decision process
}

\DeclareAcronym{LSTM}{
    short = LSTM,
    long = long short-term memory
}

\DeclareAcronym{PPO}{
    short = PPO,
    long = proximal policy optimization
}

\begin{document}

\title{
\vspace{-3.5em}\footnotesize This work has been accepted for publication at the \\
IEEE/RSJ International Conference on Intelligent Robots and Systems (IROS), 2024 
\\\vspace{3.25em}%

\LARGE \bf
Socially Integrated Navigation: \\
A Social Acting Robot with Deep Reinforcement Learning
}

\author{Daniel Flögel$^{1}$, Lars Fischer$^{1}$, Thomas Rudolf$^{1}$, Tobias Schürmann$^{1}$, and Sören Hohmann$^{2}$
\thanks{$^{1}$ are with FZI Research Center for Information Technology, Karlsruhe, Germany
        {\tt\small floegel@fzi.de}}%
\thanks{$^{2}$ is with the Institute of Control Systems at Karlsruhe Institute of Technology, Karlsruhe, Germany
        {\tt\small soeren.hohmann@kit.edu}}%
}

\maketitle
\thispagestyle{empty}
\pagestyle{empty}

\begin{abstract}
Mobile robots are being used on a large scale in various crowded situations and become part of our society.
The socially acceptable navigation behavior of a mobile robot with individual human consideration is an essential requirement for scalable applications and human acceptance.
\ac{drl} approaches are recently used to learn a robot's navigation policy and to model the complex interactions between robots and humans.
We propose to divide existing \ac{drl}-based navigation approaches based on the robot's exhibited social behavior and distinguish between social collision avoidance with a lack of social behavior and socially aware approaches with explicit predefined social behavior.
In addition, we propose a novel \textit{socially integrated navigation} approach where the robot's social behavior is adaptive and emerges from the interaction with humans. 
The formulation of our approach is derived from a sociological definition, which states that \textit{social acting} is oriented toward the acting of others.
The \ac{drl} policy is trained in an environment where other agents interact socially integrated and reward the robot's behavior individually. 
The simulation results indicate that the proposed socially integrated navigation approach outperforms a socially aware approach in terms of ego navigation performance while significantly reducing the negative impact on all agents within the environment.
\end{abstract}

\section{INTRODUCTION}

With the increasing development of smart cities, autonomous mobile robots are being integrated into our daily lives and deployed in various public pedestrian-rich environments \cite{CavalloEmotionModellingforSocialRoboticsAp2018, AlaoUncertaintyawareNavigationinCrowded2022, VargaCooperativeDecisionMakinginSharedSp2023 ,GenevoisInteractionawarePredictiveCollisionD2023}.
As the robot performs its task, it navigates through a variety of scenarios, expecting to embody the respective human social behaviors \cite{CavalloEmotionModellingforSocialRoboticsAp2018, GononInverseReinforcementLearningofPedest2023}.
Humans follow social norms, e.g., keeping a distance from others, when navigating in crowded environments \cite{YufanSociallyAwareMotionPlanningwithDeep2017, RiosMartinezFromProxemicsTheorytoSociallyAware2015}.
However, they adapt their social behavior, e.g., the exact distance to others or the velocity, to the corresponding situation, and to that of the humans around them.
The individual interpretation of appropriate social behavior is indirectly communicated to others through their movements and through social cues \cite{RiosMartinezFromProxemicsTheorytoSociallyAware2015}.

\begin{figure}[!t]
    \centering 
    \includegraphics[width=0.95\linewidth]{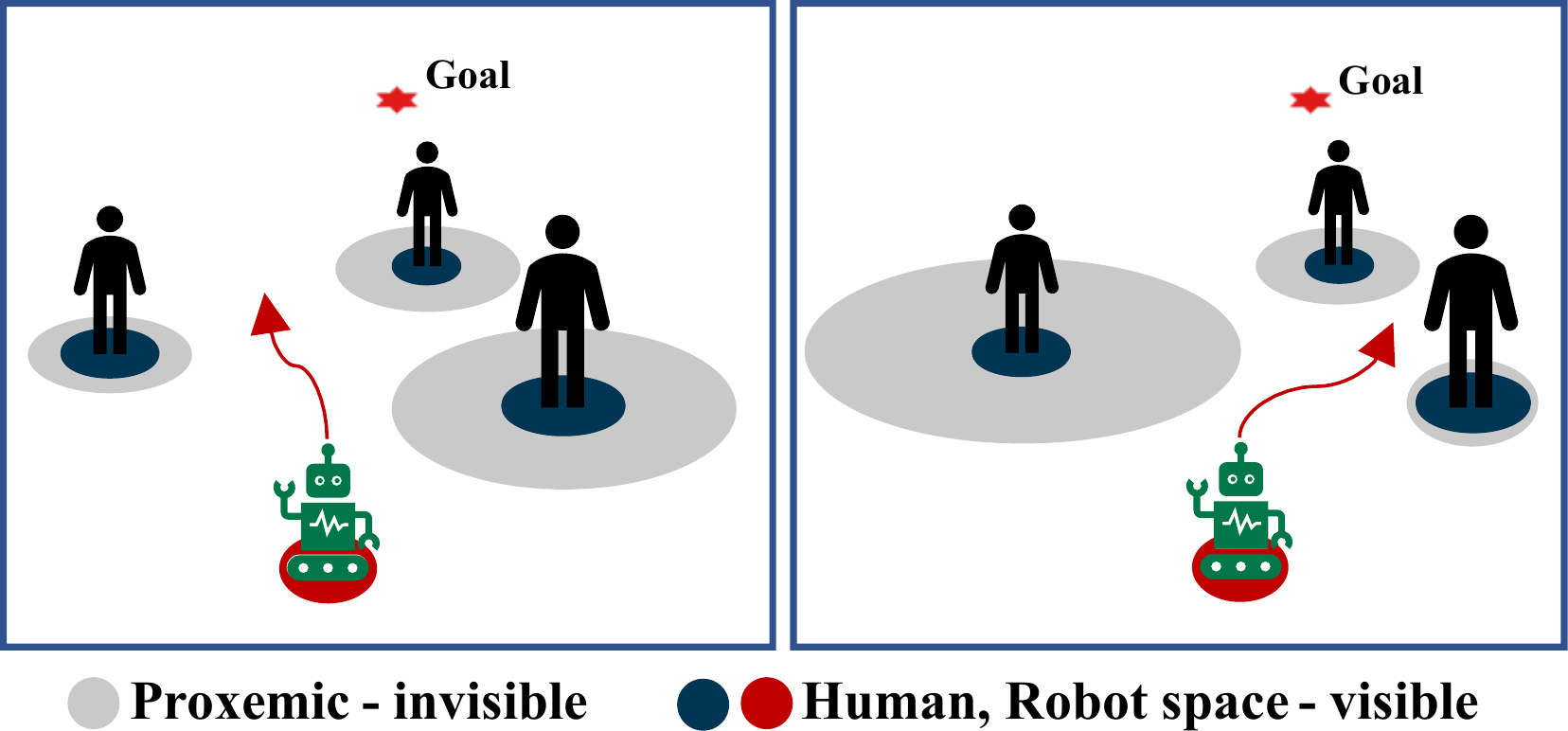}
    \caption{The proposed \textit{socially integrated navigation} approach is adaptive to human behavior and preferences. 
    It is based on a perspective change where the social behavior of the robot arises from its interaction with humans and is not predefined.
    The \ac{drl} policy has learned from the interaction behavior to consider the unknown personal space of each human. 
    }
    \label{fig:paper_summary}
\end{figure}

The navigation of a robot in crowded environments is generally referred to as social navigation and is performed by a social robot \cite{FongAsurveyofsociallyinteractiverobots2003,MavrogiannisCoreChallengesofSocialRobotNavigati2021, ZhuCollisionAvoidanceAmongDenseHeteroge2023}.
A socially intelligent robot has the same social abilities and skills as a human \cite{DautenhahnSociallyintelligentrobotsdimensions2007}. 
This field of research is, therefore, interdisciplinary and requires insights from engineering, human-robot cognition, psychology, design research, and sociology \cite{MavrogiannisCoreChallengesofSocialRobotNavigati2021, ZhuDeepreinforcementlearningbasedmobile2021}. 
This paper regards the social navigation problem from the sociological definition of \textit{social acting} where acting is termed as social if it is related to the acting of others \cite{EinfuhrunginHauptbegriffederSoziolog2016}. 
In addition, we consider the definition of \textit{socially acceptable} behavior whereby the human-machine interaction is based on social norms, the individual nature of the human is taken into account, and the robot is able to adapt to human preferences as depicted in Fig.~\ref{fig:paper_summary} \cite{DautenhahnSociallyintelligentrobotsdimensions2007}.

Social navigation approaches are classified as local motion planning \cite{YufanSociallyAwareMotionPlanningwithDeep2017, MavrogiannisCoreChallengesofSocialRobotNavigati2021} and can be divided into decoupled and coupled approaches \cite{MavrogiannisCoreChallengesofSocialRobotNavigati2021}. 
Decoupled approaches first predict human motions and then plan collision-free trajectories \cite{VargaCooperativeDecisionMakinginSharedSp2023}.
However, due to the high uncertainty in human motions, this can lead to the freezing robot problem where no collision-free trajectory can be found \cite{TrautmanUnfreezingtherobotNavigationindens2010}.
In coupled approaches, the future crowd evolution is considered as joint sequential decision-making.
Coupled approaches either assume a given structure of the problem as in game theory \cite{GalatiGametheoreticaltrajectoryplanningenh2022}, the \ac{SFM} \cite{HelbingSocialforcemodelforpedestriandynami1995}, and in \ac{ORCA} \cite{vandenBergReciprocalnBodyCollisionAvoidance2011} or assume insights about the principles as in \acf{drl} \cite{ChenDecentralizednoncommunicatingmultiage2017} or inverse reinforcement learning approaches \cite{KimSociallyAdaptivePathPlanninginHuman2016,GononInverseReinforcementLearningofPedest2023}. 

\begin{figure*}[!ht]
    \centering
    \includegraphics[width=0.99\linewidth]{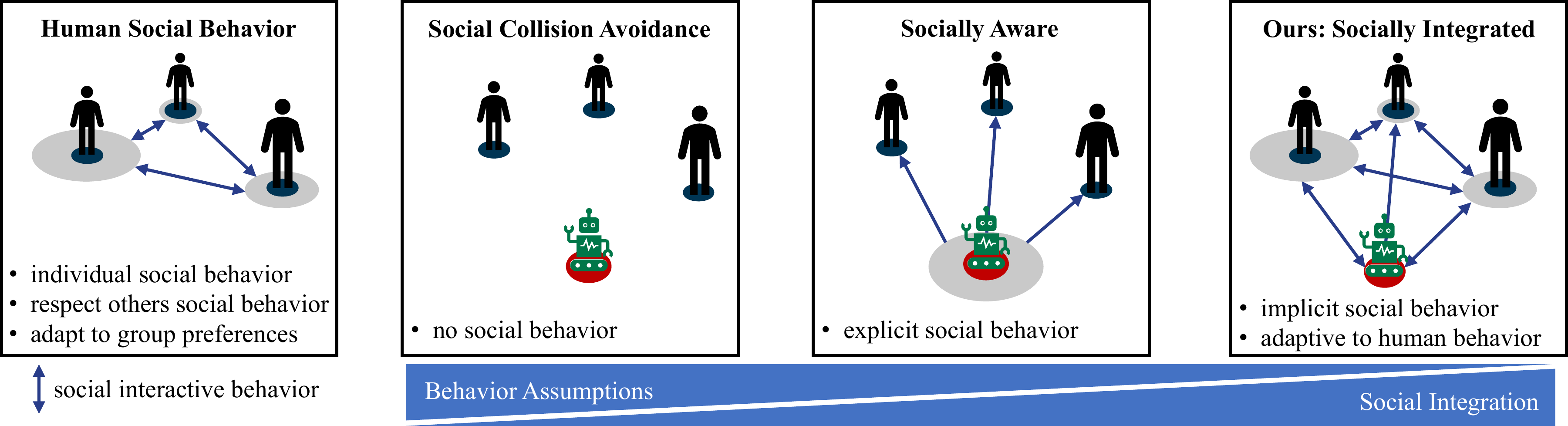}
    \caption{We propose to distinguish between different \ac{drl}-based navigation approaches among pedestrians based on robot's exhibited social interactive behavior. Social collision avoidance approaches consider only collision-free navigation.
    In socially aware approaches, a predetermined social behavior is projected onto all humans in the environment.
    Adaptive social behavior is considered in our socially integrated approaches, where humans' individual social behavior is considered, and the robot's behavior arises from an interaction with humans.}
    \label{fig:Taxonomy}
\end{figure*}

\ac{drl} approaches learn a navigation policy that maximizes the expected return over one episode \cite{ChenDecentralizednoncommunicatingmultiage2017}.
Existing approaches either only show efficient navigation behavior while avoiding physical collisions \cite{ChenDecentralizednoncommunicatingmultiage2017, FanCrowdMoveAutonomousMaplessNavigation2018, KastnerAutonomousNavigationinComplexEnviron2021,BritoWheretogonext:LearningaSubgoalRec2021} or additionally exhibit a fixed social behavior, e.g., keeping a predefined distance to all humans \cite{EverettMotionPlanningAmongDynamicDecision2019, ChanganChenCrowdRobotInteraction:CrowdawareRob2019, EverettCollisionAvoidanceinPedestrianRichE2021, ChenRelationalGraphLearningforCrowdNavi2019} as illustrated in Fig.~\ref{fig:Taxonomy}.
The robot's social behavior remains constant until retraining and every human in every scenario is treated equally regardless of how they behave. 
This contradicts socially acceptable behavior. 
These approaches exhibit no adaptive behavior, which is a requirement for social behavior from a sociological perspective \cite{EinfuhrunginHauptbegriffederSoziolog2016}.
In addition, a non-adaptive approach does not scale to different scenarios and human behavior.

We address the above-mentioned drawbacks with a \ac{drl}-based \textit{socially integrated navigation} approach, which adapts to the individual social behavior of humans in the current situation. 
The approach is derived from the sociological understanding of social acting and leads to socially acceptable robot behavior. 
We propose to rethink the navigation approaches and change the perspective of how the robot's social behavior emerges. 
Moving away from an ego-robot perspective with fixed predefined social behavior toward mutual recognition as part of the crowd, where social behavior emerges through adaptation and the basic understanding of social norms. 
The socially integrated navigation formulation enables the robot to integrate itself into the crowd through its actions.

The main contributions of this paper are (i) a novel categorization of existing \ac{drl}-based social navigation approaches based on robot's social behavior.
(ii) We propose a \ac{drl}-based socially integrated navigation approach that empowers the robot to adapt its social behavior individually to human behavior and (iii) leverage principles of training a socially integrated policy from scratch.
(iv) The evaluation with an impact analysis of the robot's behavior in the crowd shows the improvements in social adaptive robot behavior.

\section{Social Navigation}

\subsection{Social Behavior in Crowded Environments}

From a sociological perspective, the term \textit{social} constitutes behavior that is related to the interaction between people and, based on Max Weber's definition, is acting termed as social acting if it is related to the acting of other people and if the actions are oriented toward those of others \cite{EinfuhrunginHauptbegriffederSoziolog2016}. 
In this context, the behavioral rules in the interaction are denoted as social norms \cite{EinfuhrunginHauptbegriffederSoziolog2016}, and the resulting social behavior emerges from reciprocal collision avoidance \cite{YufanSociallyAwareMotionPlanningwithDeep2017}. 
This implies that social behavior arises from an interaction between the behaviors of the individual participants and is not predetermined. 
In this regard, the interpersonal communication during the interaction is non-verbal through wordless cues, referred to as social cues, that are sent and received by humans \cite{RiosMartinezFromProxemicsTheorytoSociallyAware2015}.
These social cues, such as facial expressions or motion patterns, control social interactions by expressing personal feelings and adapting individually to the reactions of others.

A human's spatial management is described by the proxemics theory and was first proposed by Hall in \cite{HallThehiddendimension1990}. 
Hall observed that people use spatial distances in various social and interpersonal interactions, whereby the distances vary depending on the environment and cultural factors.
The spatial management is divided into the intimate, personal, social, and public zones, with increasing distance to the human. 
The dynamic and situation-dependent personal space is actively maintained by humans, and the intrusion of another human causes discomfort.
Accordingly, humans try to maintain spaces for themselves that are preferred by themselves as well as by others \cite{RiosMartinezFromProxemicsTheorytoSociallyAware2015}. 

For a robot to exhibit socially acceptable behavior, it must not only be guided by social norms but also view humans as individuals and adapt to their individual preferences \cite{DautenhahnSociallyintelligentrobotsdimensions2007}.
The adaptation to context-specific social expectations, cultural norms, and individual preferences is a success factor for large-scale use of mobile robots in crowded environments \cite{DautenhahnSociallyintelligentrobotsdimensions2007, RiosMartinezFromProxemicsTheorytoSociallyAware2015, ChristoforosI.MavrogiannisValtsBlukisandRossA.KnepperSociallyCompetentNavigationPlanningb2017, RossiTheSecretLifeofRobotsPerspectives2020}. 

In summary, we conclude that a robot should have a basic understanding of social norms, but its social behavior should result from the individual interaction with a human through adaptation. 
In this way, a robot acts socially, has socially accepted behavior, and is applicable to large-scale use.

\subsection{Proposed Categorization for DRL Navigation Approaches}
We propose to distinguish \ac{drl}-based navigation approaches in crowds based on the robot's exhibited social behavior in human-machine interaction, as illustrated in Fig.~\ref{fig:Taxonomy}.
The social behavior is the basis for socially acceptable behavior as well as social acting and thus plays a fundamental role in the integration of robots into our society \cite{CavalloEmotionModellingforSocialRoboticsAp2018}.
Recent reviews of social navigation have mainly focused on general methodological categorization and whether approaches demonstrate social behavior in principle \cite{GuillenRuizEvolutionofSociallyAwareRobotNaviga2023, SingamaneniAsurveyonsociallyawarerobotnavigat2024}. 
However, they do not directly distinguish for \ac{drl}-based approaches how they exhibit social behavior and how they have learned social behavior.
Thus, we distinguish between \textit{Social Collision Avoidance}, \textit{Socially Aware Navigation}, and propose the novel category \textit{Socially Integrated Navigation} as described in the following. 

\subsubsection{Social Collision Avoidance}
We refer to a robot navigation policy that is trained to efficiently reach the goal position while avoiding physical collisions in an environment with dynamic agents as social collision avoidance. 
Thereby, agents in the environment are referred to as humans but neither the robot nor the agents consider social norms or social behavior. 
The social aspect of these approaches arises from efficient and collision-free navigation next to humans. 
During the training procedure, the agent is rewarded for reaching the goal, penalized for colliding with other agents, and encouraged to efficient navigation \cite{FanCrowdMoveAutonomousMaplessNavigation2018,BritoWheretogonext:LearningaSubgoalRec2021, KastnerAutonomousNavigationinComplexEnviron2021, AhSenHumanAwareSubgoalGenerationinCrowde2022}.

\subsubsection{Socially Aware Navigation}
We refer to a navigation policy that is, additionally to social collision avoidance, trained to respect explicit social behavior as socially aware navigation. 
Accordingly, the robot is rewarded for maintaining a predefined social behavior.
This social behavior remains constant, and the robot interacts with every human in the same way, regardless of how the human behaves, as depicted in Fig.~\ref{fig:Taxonomy}.

Training a robot to maintain a minimum distance from humans was first proposed with CADRL in \cite{ChenDecentralizednoncommunicatingmultiage2017} and augmented to an arbitrary number of humans in \cite{EverettMotionPlanningAmongDynamicDecision2019, EverettCollisionAvoidanceinPedestrianRichE2021} using a \ac{LSTM} to encode the human-robot interaction.
This pairwise interaction was augmented to attention-based crowd-robot interaction in \cite{ChanganChenCrowdRobotInteraction:CrowdawareRob2019},  reformulated as a directed graph in \cite{ChenRelationalGraphLearningforCrowdNavi2019} and augmented with an attention mechanism in \cite{ChenRobotNavigationinCrowdsbyGraphConv2019, ZhouLearningCrowdBehaviorsinNavigationw2024} to avoid performance degradation in large crowds.  
These approaches focus on efficient crowd-robot interaction modeling.
Social behavior is restricted to maintaining a constant minimum distance from other humans, which can be seen as the robot's personal zone. 

Velocity-depending danger zones, which are differently shaped personal zones, were proposed in \cite{SamsaniSociallyCompliantRobotNavigationinC2021} and augmented with a hazardous area in \cite{XueCrowdAwareSociallyCompliantRobotNav2023}. 
Oriented bounding capsules were proposed in \cite{ZhuCollisionAvoidanceAmongDenseHeteroge2023} to model non-circular shaped agents and velocity-depending risk-zones in moving direction.
Emotional depending shaped personal spaces are proposed in \cite{NarayananEWareNet:EmotionAwarePedestrianInten2023} with a predefined set of comfort spaces.
A risk-map-based approach with human position prediction is proposed in \cite{YangRMRLRobotNavigationinCrowdEnvironm2023} to represent crowd interactions and geometric structures.
These approaches consider variable personal zones but with predefined fixed assumptions from an ego-robot perspective.

A norm-inducing reward function is proposed in \cite{YufanSociallyAwareMotionPlanningwithDeep2017} to train an agent with overtaking, passing, and crossing capability. 
However, the agent is explicitly trained to do that in a predefined manner, e.g., passing on the right, and deviations from the predetermined social behavior are due to the tradeoff to time optimality and not due to adaptivity to human preferences. 
Hence, this is a predefined social behavior rather than an adaptation to human preferences.

All these approaches are based on predefined social behavior and show no adaptivity to human behavior. 
As a result, these approaches are not socially acceptable based on the considered definitions \cite{DautenhahnSociallyintelligentrobotsdimensions2007, EinfuhrunginHauptbegriffederSoziolog2016}. 
In addition, if the prior assumptions for social behavior do not apply, these approaches are not efficiently applicable.

\subsubsection{Socially Integrated Navigation}
We refer to a navigation policy as socially integrated, which is trained to exhibit implicit and adaptive social behavior to other agents' behavior with individual consideration. 
Adaptation is not an imitation but rather a deviation from a personal understanding.
In addition, the policy must be trained in an environment where other agents exhibit social behavior as well. 
There is no existing \ac{drl} approach that fulfills these minimum requirements toward a socially integrated navigation.

\section{Preliminaries}

Throughout this paper, a dynamic object in the environment is generally referred to as an agent, either a robot or a human, and its behavior is determined by a policy.
Variables referred to the robot are indexed with $x_0$, and humans with $x_i$ with $i \in 1, \cdots N-1$. 
A scalar value is denoted by $x$ and a vector by $\bm{x}$.

\subsection{Problem Formulation}
The navigation task of a robot toward a goal in a pedestrian-rich environment is a sequential decision-making problem and can be modeled as \ac{POMDP} \cite{ChenDecentralizednoncommunicatingmultiage2017, ChanganChenCrowdRobotInteraction:CrowdawareRob2019, ZhuDeepreinforcementlearningbasedmobile2021}. 
The \ac{POMDP} is described with a 7-tuple $(\mathcal{S}, \mathcal{A}, \mathcal{T}, \mathcal{O}, \mathcal{T}_0, R, \gamma)$.
We assume the state space $\mathcal{S}$ and action space $\mathcal{A}$ as continuous.
The transition function $\mathcal{T} : \mathcal{S} \times \mathcal{A} \times \mathcal{S} \rightarrow [0,1]$ describes the probability transitioning from state $\bm{s}_{t} \in \mathcal{S}$ to state $\bm{s}_{t+1} \in \mathcal{S}$ for the given action $\bm{a}_t \in \mathcal{A}$. 
With each transition, an observation $\bm{o}_t \in \mathcal{O}$ and a reward $R : \mathcal{S} \times \mathcal{A} \rightarrow \mathbb{R}$ is returned by the environment. 
The initial state distribution is denoted by $\mathcal{T}_0$ while $\gamma \in [0,1)$ describes the discount factor.
The \ac{POMDP} serves as a framework for the problem formulation and is solved with a \ac{drl} approach \cite{RichardSuttonReinforcementLearning2020}. 
The environment contains $N$ agents with one robot ($i=0$) and $N-1$ humans ($i \in {1, \ldots, N-1}$).
In single robot navigation, only the robot's navigation policy $\pi_0$ is trained.
The human policies $\pi_i$ are unknown to the robot and correspond to an existing behavioral model. 
Every agent is completely described with a state $\bm{s}_{i, t} = [\bm{s}_{i,t}^o , \bm{s}_{i,t}^h]$ at any given time $t$, where the state is separable into an observable $\bm{s}_{i,t}^o$ and unobservable $\bm{s}_{i,t}^h$ part:
\begin{align} 
    \bm{s}_i^o &= 
    \begin{bmatrix}
        \bm{p},& \bm{v},& r
    \end{bmatrix}
     \in \mathbb{R}^5 \label{eq:state_observable}\\
    \bm{s}_i^h &= 
    \begin{bmatrix}
        \bm{p}_g,& v_{\text{pref}},& \psi_{\text{pref}},& r_{\text{prox}}
    \end{bmatrix}
     \in \mathbb{R}^5 \label{eq:state_hidden}
\end{align}
The observable state is known to everyone and composed of the position $\bm{p} = [p_x, p_y]$, velocity $\bm{v}=[v_x, v_y]$, and radius $r$. 
We assume agents as circular objects where the radius determines the area which is occupied by this agent. 
This assumption is valid since we consider an environment with only humans and a robot with a human-like shape.
For agents with other shapes, oriented bounding capsules are proposed in \cite{ZhuCollisionAvoidanceAmongDenseHeteroge2023}.
The unobservable state is only known to the agent itself and composed of goal position $\bm{p}_g = [p_{gx}, p_{gy}]$, preferred velocity $v_{\text{pref}}$, preferred orientation $\psi_{\text{pref}}$, and a proxemic radius $r_{\text{prox}}$. 
The proxemic radius represents the agent's personal space according to Hall's proxemic theory.  
The world state $\bm{s}_t = [\bm{s}_{0,t}, \cdots, \bm{s}_{N,t}]$ represents the current situation in the environment.

We regard the \ac{POMDP} as observable given sufficient partial history, whereby the hidden states of an agent can be inferred from past observations.
In every timestep $t$, the robot observes the environment receiving $\bm{o}_t$ and takes an action $\bm{a}_t = \pi(\bm{a}_t|\bm{o}_t)$. 
The robot kinematic is assumed to be a unicycle model and the action $\bm{a}_t = [v_t, \Delta \theta_t]$ is the velocity and heading change, respectively. 
The trajectory $\tau$ of one episode is the sequence of states, observations, actions, and rewards within the terminal time $T$.
The return of one episode $\mathcal{R}(\tau) = \sum_{t=0}^T \gamma^t R_t$ is the accumulated discounted reward $R_t$ and the central objective is to learn the optimal robot policy $\pi^*$ which maximizes the expected return:
\begin{align}
    \mathcal{T}(\tau|\pi) &= \mathcal{T}(\bm{s}_0) \prod_{t=0}^T \mathcal{T}(\bm{s}_{t+1}|\bm{s}_t, \bm{a}_t) \pi(\bm{a}_t|\bm{o}_t)\\
    \underset{\tau \sim \pi}{\mathbb{E}} [\mathcal{R}(\tau)] &= \int_\tau \mathcal{T}(\tau|\pi)R(\tau)\\
    \pi^*(\bm{a}|\bm{o}) &=  \arg \underset{\pi}{\max} \underset{\tau \sim \pi}{\mathbb{E}} [\mathcal{R}(\tau)]
\end{align}
Considering a stochastic environment, $\mathcal{T}(\tau|\pi)$ is the probability of a trajectory starting in $\bm{s}_0$ with the probability $\mathcal{T}_0$.

\section{Approach}
This section presents our \ac{drl}-based socially integrated navigation approach, which is based on social acting and learns to adapt to the social behavior of humans. 
We first describe how we incorporated the social integration formulation within a \ac{drl} framework, and subsequently, we impose minimum requirements on the training process. 

\subsection{Adapting Social Behavior}

Recall that the objective of the \ac{drl} training process is to find a policy that maximizes the expected accumulated discounted reward over one episode. 
The reward system is decisive for what the policy has learned from the observation, whereas the observation must be sufficient to learn the desired behavior.
In the environment, humans have their own objectives and evaluate the current situation and the behavior of others based on their personal preferences and objectives. 
This assessment is then transferred to other people in the environment through social cues.

\begin{figure}[!t]
    \centering
    \includegraphics[width=0.8\linewidth]{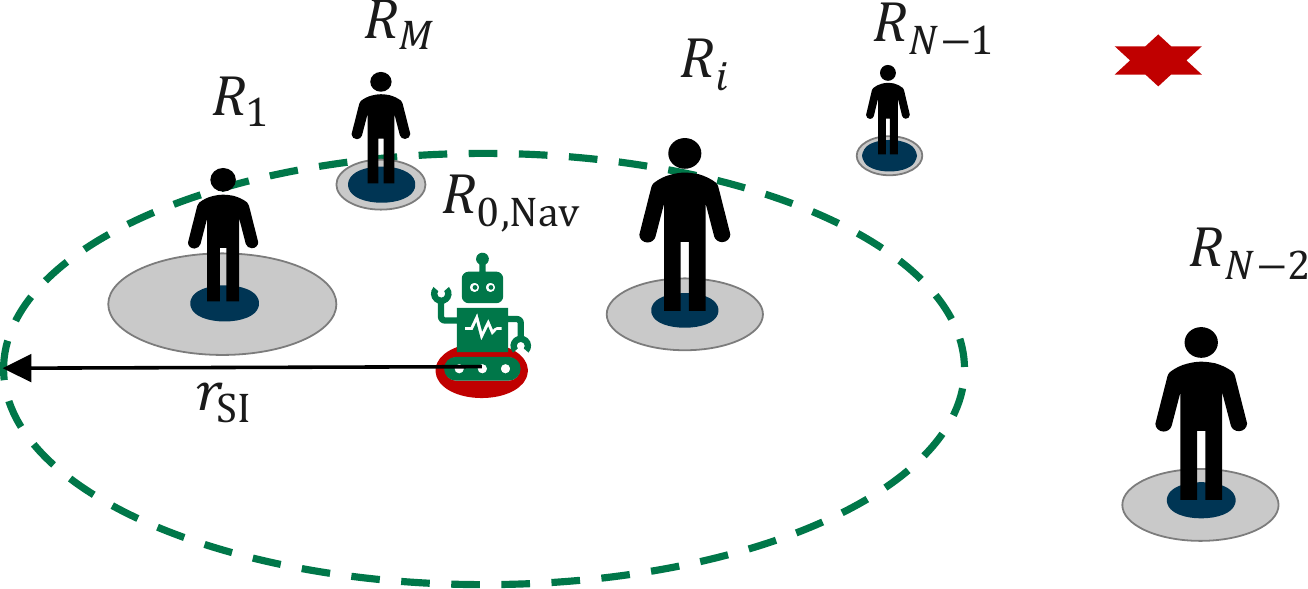}
    \caption{Proposed distributed rewards in the environment. The robot has only one navigation reward and receives the reward for the social behavior of humans within the social integration radius $r_{\text{SI}}$.}
    \label{fig:SI-Informtaionsebene}
\end{figure}

This leads to an environment where multiple reward systems exist, as depicted in Fig. \ref{fig:SI-Informtaionsebene}.
We interpret the social cues as a reward a human will give the robot. 
We assume an integrated society where everyone is perceived as part of the group, and therefore, we assume that people treat the robot equally to humans. 
This individual human reward evaluates the robot's behavior from the individual human perspective, and all rewards together evaluate the robot's behavior in the group. 
Since not every human in the environment gives social cues, e.g., because the distance is too large or he is not directly involved in the interaction, only the people within a certain radius $r_{\text{SI}}$ around the robot are considered as illustrated in Fig. \ref{fig:SI-Informtaionsebene}. 
In addition, we consider the adaptation of social behavior as a local process. 
From this principle, we derive the socially adaptive reward
\begin{align}
    R_{\text{SA}} = \frac{1}{M} \sum_{i=1}^M \lambda_i R_i
\end{align}
that the robot receives at each time step.
With the reward $R_i$ of human $i$, a scaling factor $\lambda_i$, and the number of humans $M$ within the social integration radius $r_{\text{SI}}$ around the center of the robot. 
If the human $i$ is within the social integration radius of the robot $d_{0i} < r_{\text{SI}}$, the human reward is given at every time step by
\begin{equation}
    R_i = 
        - R_v \cdot |v_{i,t} -v_{0,t}| -  R_{\text{prox}} .
    \label{eq:Reward_SA}
\end{equation}
The reward comprises velocity deviations scaled with $R_v$ and rewards violations of humans' personal space $d_{i0} < r_{i,\text{prox}}$ with $R_{\text{prox}}$.
We observed that using sparse reward for proxemic violations leads to better results and generalization than dense reward. 
The scaling factor allows the robot to interpret and evaluate the received reward and to pay more attention to certain humans.
For example, more focus can be paid to people who are close or further away from the robot. 
The social integration radius is decisive for how granularly local group situations are regarded.
Note that being socially integrated does not mean imitating others.
The robot forms its understanding and interpretation of the current situation from its perspective and adapts to the group behavior, taking people individually into account. 
In addition to the social adaptive reward, the robot is rewarded for efficient navigation with $R_{\text{Nav}}$ which leads to the reward at time step $t$
\begin{align}
    R_t = R_{\text{Nav}} + R_{\text{SA}} .
\end{align}
The navigation reward is responsible for teaching the robot efficient navigation toward the goal while avoiding collisions.
\begin{equation}
R_{\text{Nav}} = \begin{cases}
    + R_g & \textrm{if} \hspace{3mm} \bm{p}_t=\bm{p_g} \\
    - R_c & \textrm{if} \hspace{3mm} d_{0i} \leq 0 \\
    - R_{\text{time}} & \textrm{if} \hspace{3mm} \text{timeout} \\
    + R_{gd,1} \cdot |\Delta d_g|  & \textrm{if} \hspace{3mm} \Delta d_g  > 0 \\
    - R_{gd,2} \cdot  |\Delta d_g|  & \textrm{if} \hspace{3mm} \Delta d_g  < 0 \\
    \end{cases}
\end{equation}
We encourage to make steps toward the goal with $R_{gd,1}$ with $\Delta d_g = d_{g,t} - d_{g,t-1}$ and reaching the goal with $R_g$. 
Contrarily, we penalize the robot for collisions with other agents with $R_c$, running into timeouts with $R_{\text{time}}$, and making steps from goal away with $R_{gd,2}$.
The dense navigation rewards support fast and stable training.


A key requirement to adapt to the social behavior of others is that the others follow their social norms, and the robot can observe their behavior.
This means that the social norms must be observable based on the observation the agent does, and the social norms must be observable based on the interaction with the other agent. 
This implies that the agent must consider the temporal crowd evolution and spatial relationships between the agents.

The observation is divided into the robot observation $\bm{o}_{0,t}$ and human observation $\bm{o}_{i,t} = [\bm{\Bar{o}}_i, \bm{\hat{o}}_{i,t}] $. 
Based on human's observable states, the robot observes the human $i$ at time $t$: 
\begin{align}
\bm{o}_{0,t} &= 
\begin{bmatrix}
    d_g & \Delta \bm{p}_g & \theta & v_{\text{pref}} & r_0 
\end{bmatrix}\\
\bm{\Bar{o}}_i &=
\begin{bmatrix}
    r_i & r_i + r_0 
\end{bmatrix} \\
\bm{\hat{o}}_{i,t} &=
\begin{bmatrix}
    d_t & \Delta \bm{p}_t & \Delta \bm{v}_t
\end{bmatrix}
\end{align}
The robot observes its state based on the relative position to the goal $\Delta \bm{p}_g = \bm{p}_g - \bm{p}_t$, the direct distance to the goal $d_g = ||\Delta \bm{p}_g||_2$, the heading $\theta$, the preferred velocity $v_{\text{pref}}$, and the personal radius $r$.
We distinguish the human observation in a constant part $\bm{\Bar{o}_i}$ which does not change over time, and a time-varying part $\bm{\hat{o}}_{i,t}$ which varies throughout the steps. 
The constant part contains the human radius $r_i$ and the combined radius $r_i + r_0$. 
The temporal observation $\bm{\hat{o}}_{i,t}$ contains the relative position $\Delta \bm{p}_t =\bm{p}_{0,t} - \bm{p}_{i,t}$, direct distance $d_t = ||\Delta \bm{p}_t||_2$, and relative velocity $\Delta \bm{v}_t = \bm{v}_{0.t} - \bm{v}_{i,t}$ between robot and human. 
To consider the partial history, we concatenate the last $k$ temporal observations to the aggregated observation $\bm{o}_{i,t}$ of human $i$ at timestep $t$
\begin{align}
    \bm{o}_{i,t} = 
    \begin{bmatrix}
        \bm{\Bar{o}}_{i,t}& \bm{\hat{o}}_{i,t}& \bm{\hat{o}}_{i,t-1} & \cdots& \bm{\hat{o}}_{i,t-k} 
    \end{bmatrix} .
\end{align}
In a human-aware perspective, the joint observation 
\begin{align}
    \bm{o}_t = 
    \begin{bmatrix}
        \bm{o}_{0,t}, \bm{o}_{1,t}\\
        \cdots \\
        \bm{o}_{0,t}, \bm{o}_{N-1,t}
    \end{bmatrix}
\end{align}
encodes the pairwise robot-human interaction capacity and temporal evolution.

\subsection{Training a Socially Integrated Agent}
The formulation of our socially integrated navigation approach based on social acting refers to the behavior of others and adapting to their behavior.
In a real-world setting, the robot is used in an environment where humans respect the social behavior of others and have learned to adapt to their behavior, and they interact socially integrated.
To ensure realistic and representative training, we impose the minimum requirement to the training environment that humans act according to their own social behavior and that of other humans.
This implies that they actively avoid violating their own preferences and those of others.
In addition, the social behavior of humans must correspond to the social norms to which the robot should adapt in the real world.
Otherwise, the robot can't infer social behavior from social cues and observations, and thus, it is not possible to train a socially integrated robot.

\begin{figure}[!tb]
    \centering
    \includegraphics[width=0.80\linewidth]{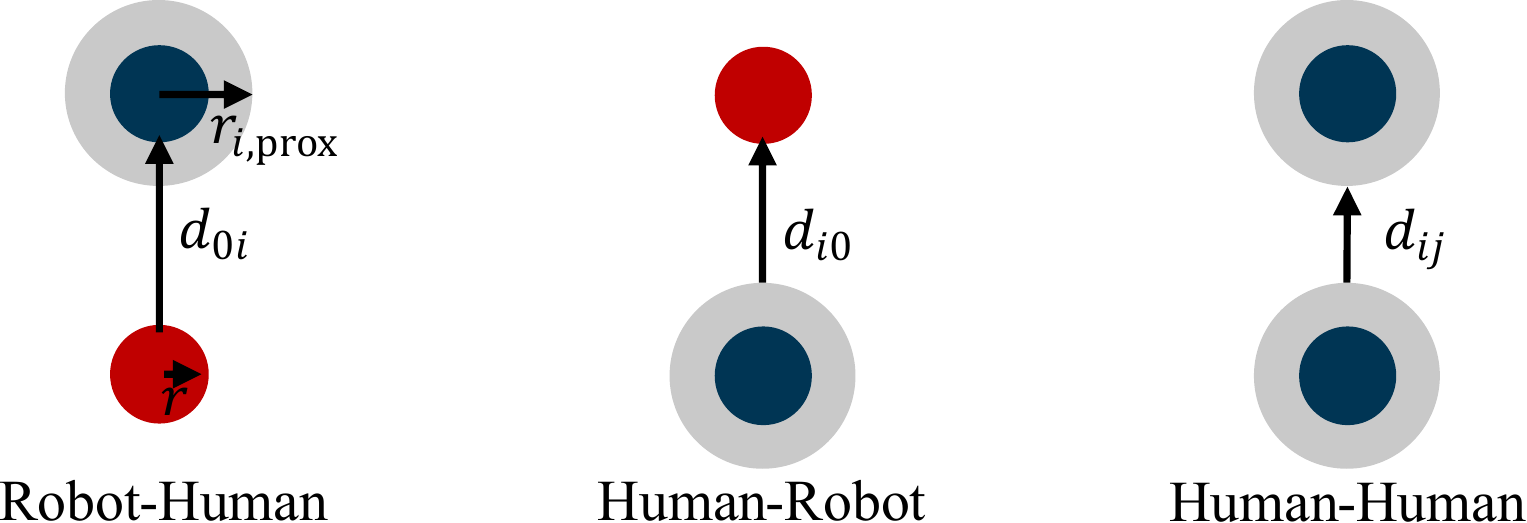}
    \caption{Proposed perspective of the various agents among each other in the environment. The gray-shaded personal areas are only visible to the respective agent. Humans act in the environment to maintain their personal space and that of others. }
    \label{fig:environment_agents}
\end{figure}

We follow the proposal to use the \ac{ORCA} model to simulate human interactive behavior \cite{ChenRelationalGraphLearningforCrowdNavi2019, EverettCollisionAvoidanceinPedestrianRichE2021, YaoCrowdAwareRobotNavigationforPedestr2021, YangRMRLRobotNavigationinCrowdEnvironm2023} since it is more stable than \ac{SFM}-based approaches in densly crowded scenarios \cite{GuillenRuizEvolutionofSociallyAwareRobotNaviga2023}.
In a nutshell, \ac{ORCA} computes a collision-free velocity for the ego robot within a time horizon, assuming that other agents in the environment follow the same policy, and thus \ac{ORCA} does not take into account direct social behavior. 
A cooperation coefficient is used to split the collision avoidance effort among all agents. 
As the robot is a technical machine, it has no personal space, only humans.
In an environment with one robot and $N-1$ humans, there are three basic perspectives: robot-human, human-robot, and human-human. 
The key difference in these perspectives is how others are perceived and how they interact with others. 
Fig.~\ref{fig:environment_agents} illustrates the perspectives for the case where each agent has its personal space and humans act socially integrated. 
Recall, according to the observable and unobservable state in \eqref{eq:state_observable} and \eqref{eq:state_hidden} of an agent, the human proxemic radius $r_{\text{prox}}$ is not within the observation.
Only the radius $r_i$ and position of each agent can be observed. 
This leads the robot to observe the distance $d_{0i}$ between the robot and human $i$ since the robot can not observe the hidden states. 
To allow the \ac{ORCA} models to behave appropriately according to individual social behavior, we give these models general knowledge about the entire environment.
Therefore, humans aim to maintain $d_{i0} \geq 0$ when interacting with the robot. 
In human-human interaction, they react according to $d_{ij} \geq 0$ and thus respect each proxemic radius. 
This implies that the robot can deduce individual human behavior from human-robot and human-human interaction observations. 

\begin{figure}[!tb]
    \centering
    \includegraphics[width=0.90\linewidth]{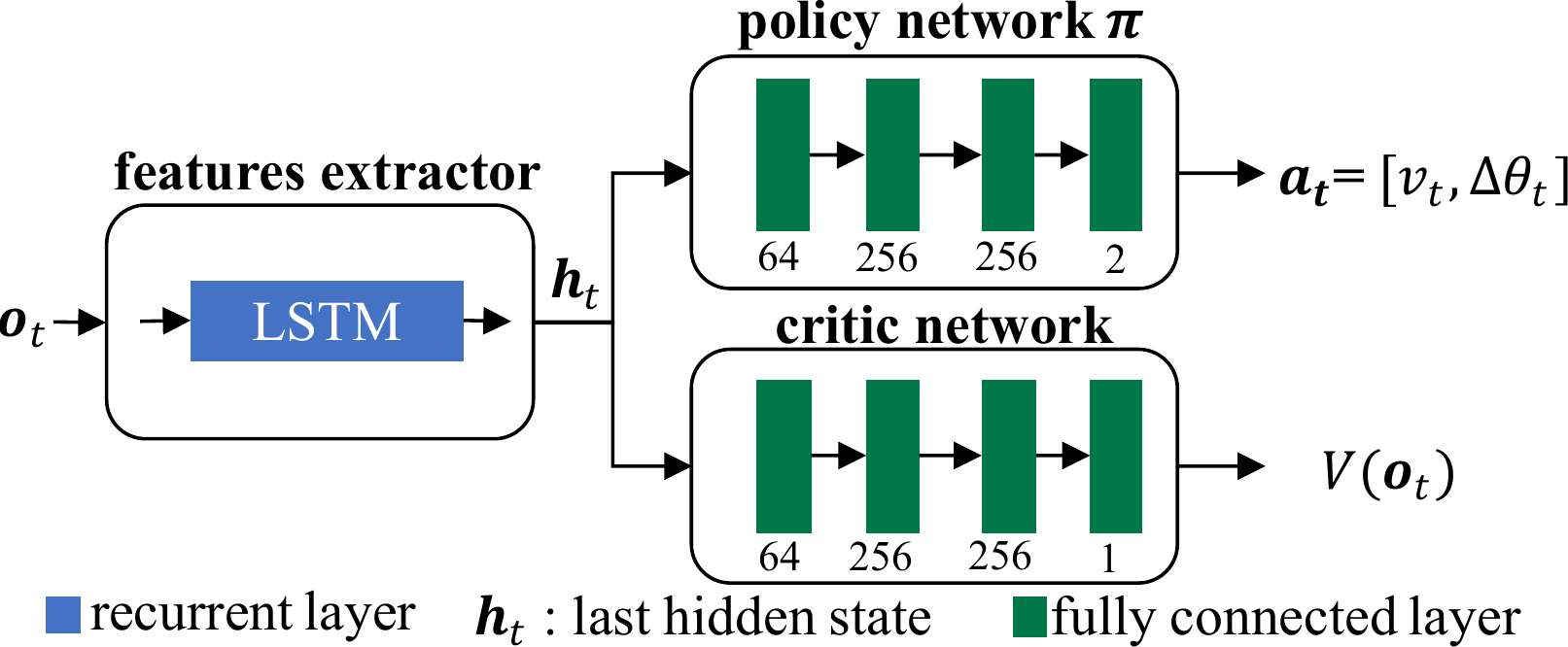}
    \caption{Proposed architecture with shared features extractor and separated policy and critic networks. The last hidden state of the \ac{LSTM} is used as input for fully connected layers.}
    \label{fig:network_architecture}
\end{figure}

The navigation policy is trained with the \ac{PPO} \cite{SchulmanProximalPolicyOptimizationAlgorithms2017} algorithm by simultaneously learning a critic network and a policy. 
We follow \cite{EverettCollisionAvoidanceinPedestrianRichE2021} to use a \ac{LSTM} as a feature extractor to handle a varying number of humans but use separated policy and critic networks. 
A comparison across $15$ seeds showed that a shared feature extractor architecture, as depicted in Fig. \ref{fig:network_architecture}, converges faster to a mean success rate of $1$ with higher reward and better navigation metrics.
Recent approaches proposed to apply imitation learning with expert trajectories in advance of the \ac{drl} training process to initialize the policy for stable and converging training \cite{ChanganChenCrowdRobotInteraction:CrowdawareRob2019, EverettCollisionAvoidanceinPedestrianRichE2021,ZhuCollisionAvoidanceAmongDenseHeteroge2023, ChenRobotNavigationinCrowdsbyGraphConv2019, ChenRelationalGraphLearningforCrowdNavi2019,BritoWheretogonext:LearningaSubgoalRec2021}.
However, imitation learning with expert knowledge introduces a bias through the expert behavior. 
Thus, we propose to reward the robot's velocity to avoid imitation learning. 
This ensures that the policy can be trained from scratch and has no bias caused by expert trajectories.
In our socially integrated formulation, a velocity reward can either be incorporated into the navigation reward $R_{\text{Nav}}$ with adding $ -R_v \cdot |v_{i,t} - v_{i,\text{pref}}| $ to train the robot having one preferred velocity or into the $R_{\text{SA}}$ to train the robot adapting to the velocity of others as denoted in \eqref{eq:Reward_SA}.

\section{Evaluation}

\begin{figure*}[!tb]
  \centering
  \subfloat[socially integrated ]{%
    \includegraphics[width=0.99\linewidth]{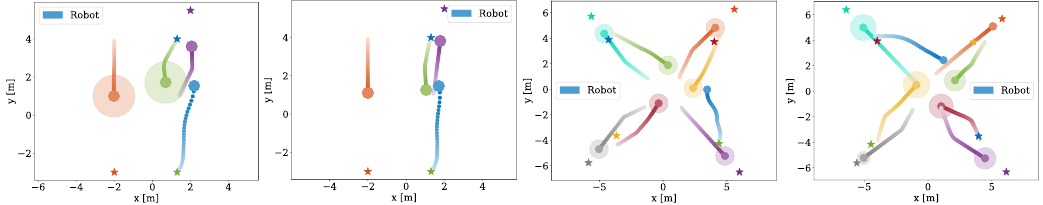}
    \label{fig:results_si}%
  }\\
  \subfloat[socially aware ]{%
    \includegraphics[width=0.99\linewidth]{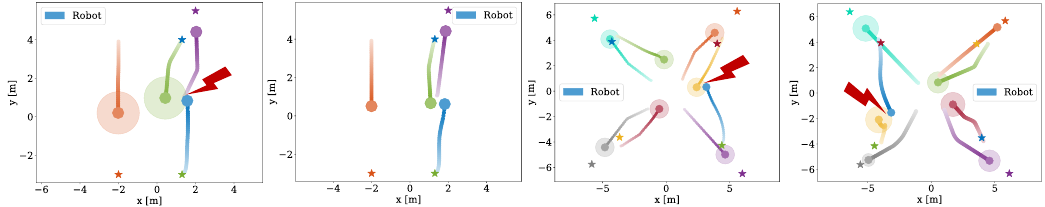}
    \label{fig:results_sa}%
  }
  \caption{Comparison of our proposed socially integrated navigation approach with adaptive social behavior in (a) to a socially aware navigation approach with fixed social behavior in (b).
  A passing scenario is depicted in column $1$ and $2$ on the left side and a homogenous and heterogenous circle crossing scenario in column $3$ and $4$ to the right. 
  Humans' personal space, depicted with faded circles around an agent, is invisible to the robot and uniformly sampled. 
  Violations of humans' personal space are marked with a red spark.}
  \label{fig:results_si_sa}
\end{figure*}

\subsection{Experimental Setup}
We consider three different scenarios for evaluation. 
One passing scenario representing a pedestrian zone, depicted in Fig. \ref{fig:results_si_sa} column $1$ and $2$, and two circle crossing scenarios representing densely crowded environments, depicted in Fig. \ref{fig:results_si_sa} column $3$ and $4$ on the right side.  
Throughout one episode, we keep the hidden state $\bm{s}_i^h$ of all agents constant, and all agents have the same radius $r_i=0.3$.
The circle crossing scenario includes $8$ agents with uniformly sampled start and end positions.  
We distinguish between a homogeneous scenario, where each human has the same random proxemic radius, and a heterogeneous scenario, where each human has its random individual proxemic radius. 
In both circle crossing scenarios, the proxemic radius $r_{\text{prox}}$ is uniformly sampled from $[0, 0.8]$, according to the intimate and personal zone in \cite{HallThehiddendimension1990}, and the human velocity is uniformly sampled between $[0.5, 1]$.
Humans are controlled by an \ac{ORCA} policy and additionally respect the social behavior of others according to the training procedure for socially integrated agents. 
The cooperation coefficient is uniformly sampled from $[0.3, 0.7]$ whereas $0.5$ means that the agent will apply half the effort to avoid a collision within the time horizon of $5$s. 
Since we aim to integrate the robot into the human group and the society, we do not evaluate our approach in scenarios where the robot is invisible to humans as done in \cite{ChanganChenCrowdRobotInteraction:CrowdawareRob2019,XueCrowdAwareSociallyCompliantRobotNav2023}.

To conduct the experiments and train the \ac{drl} policy, we developed a customized gymnasium environment \cite{TowersGymnasium2023} inspired by \cite{EverettCollisionAvoidanceinPedestrianRichE2021} and incorporated the \ac{drl} framework \textit{Stables Baselines3} \cite{AntoninRaffinStableBaselines3ReliableReinforcemen2021}.
We compare our socially integrated agent against a socially aware agent with explicit social behavior considering spatial observation ($k=0$) according to \cite{EverettMotionPlanningAmongDynamicDecision2019, EverettCollisionAvoidanceinPedestrianRichE2021, ChanganChenCrowdRobotInteraction:CrowdawareRob2019, ChenRelationalGraphLearningforCrowdNavi2019, ChenRobotNavigationinCrowdsbyGraphConv2019} and temporal observation ($k=15$).
The socially aware agent has the same policy architecture and navigation reward $R_{\textit{Nav}}$ but the robot's social behavior emerges through an ego perspective of \eqref{eq:Reward_SA} with $r_{0, \text{prox}} = 0.2$ to maintain a minimum distance to all humans and a predefined velocity.

\subsection{Results}
\subsubsection{Avoiding Imitation Learning}
We trained the socially integrated and socially aware policy over $15$ seeds, using the hyperparameters and rewards from Table \ref{tab:experiment_parameters}, and observed convergence after $1.8$\,M and $2$\,M steps, respectively, across all seeds. 
The temporal observation includes the human motions of the last $3$\,s, leading to $k=15$ with a discrete step size of $\Delta t=0.2$\,s. 
To avoid imitation learning with expert trajectories, we investigate the impact of the introduced velocity reward on the training behavior.
We train the policy with ($R_v=0.052$) and without ($R_v=0$) velocity reward in a circle crossing scenario over $15$ seeds, each with a maximum of $10$\,M steps.
The navigation performance of the trained policies reveals that using a velocity reward in training leads to a navigation policy with task completion in all $15$ seeds, whereas without the velocity reward, only $1$ trained policy could solve the navigation task. 
\begin{table}[!tb]
    \centering
    \caption{PPO Hyperparameters and Rewards }
    \vspace{-0.5em}
    \label{tab:experiment_parameters}
    \begin{tabular}{l c || l c}
        \hline
        Hyperparameter & Value & Reward & Value \\
        \hline
        optimizer & \texttt{Adam}        & $R_g$ &  4            \\
        learning rate & \num{3e-4}       &  $R_c$   &4  \\
        environment steps & \num{2048}   &   $R_{\text{time}}$ &4   \\
        clip range & 0.2                  &   $R_{gd,1}$ & 0.1 \\
        number of epochs & \num{2}           &   $R_{gd,2}$ & 0.2   \\
        batch size & \num{64}              &   $R_v$ & 0.052  \\
        discount factor $\gamma$ & \num{0.99} &  $R_{\text{prox}}$& 1.1 \\
        activation functions & \texttt{ReLU}           &   \\
        \hline
    \end{tabular}
    \vspace{-2em}
\end{table}

\subsubsection{Adapting Social Behavior}
To prove the adaptivity of our approach to individual human behavior and the impact on others, we consider the same passing scenario twice as depicted in Fig. \ref{fig:results_si_sa} on the left half.
In column $1$, the approaching other agents have an invisible personal space of $r_{\text{prox}}=0.8$, and in column $2$, they have no personal space. 
The socially integrated robot successfully adapts the distance to the approaching agents depending on the humans' personal space, whereas the socially aware agent violates the other humans' personal space while moving straight toward the goal. 
Although the human acts to maintain his personal space. 
In contrast, the socially integrated agent accelerates so that he passes the green agent earlier when he is already interacting with the purple agent. 
If the passing agents have no personal space, both robots perform similarly. 
In the heterogeneous scenario in column $4$ of Fig. \ref{fig:results_si_sa}, the socially aware robot violates the personal space of the yellow agent and also pushes him back as the yellow agent acts to maintain its personal space. 
In contrast, the socially integrated robot passes the other agents on the other side and thus avoids the conflict. 

The ablation study, based on the heterogeneous (HE) and homogeneous (HO) circle crossing scenarios, is aggregated in Table \ref{tab:quantitative_results} for a socially integrated and a socially aware robot with history (k=15) and without (k=0). 
The two ratios indicate how much more or less time and distance is required compared to the ideal value.
The human return value aggregates \eqref{eq:Reward_SA}.
The socially aware robots cause more collisions, run into more timeouts, and cause significantly more proxemic violations.
A socially aware robot with history has improved ego navigation performance but more negative impact on humans, measured by more negative human return, more proxemic violations, and worse human time ratio.
In comparison, the socially integrated robot has better ego navigation performance, causes less human discomfort, reflected in human return and proxemic violations, and leads humans to slightly better distance and time ratios.

\subsection{Discussion}
Through the extension of the \ac{ORCA} model with individual proxemic radius and the environment design that humans respect the proxemics of others, the simulated human social behavior is closer to the reality than just using the ORCA model as used in previous approaches.
Therefore, the environment is well suited to investigate the effect of the robot's exhibited social behavior.
For the socially aware navigation, the results clearly show a trade-off between an improvement in robot navigation performance and causing negative effects on other agents in the environment. 
At the same time, the socially aware agents do not react to people's conscious actions and push them back, for example. 
In contrast, the socially integrated approach does not have this trade-off. 
The perspective change leads to a significantly reduced negative impact on others and at the same time to better robot navigation performance.
In addition, due to its ability to adapt to the individual behavior of others, the socially integrated robot can react more consciously to the reactions of others and thus make more far-sighted decisions. 
However, if others are not cooperative and do not interact, the robot can not adapt, and proxemic violations can still occur.
Overall, a socially integrated robot outperforms a socially aware robot. 

\begin{table*}[!tb]
\centering
\caption{Ablation study: Our socially integrated and a socially aware (SA) agent with and without history in $100$ random episodes.}
\label{tab:quantitative_results}
\begin{tabular}{|c|c|c|c|c|cc|cc|c|}
\hline
\multicolumn{1}{|c|}{\multirow{2}{*}{Scenario}}   & \multicolumn{1}{c|}{\multirow{2}{*}{Method}} & \multicolumn{1}{c|}{\multirow{2}{*}{Col.}} & \multicolumn{1}{c|}{\multirow{2}{*}{\begin{tabular}[c]{@{}c@{}}Time- \\ out\end{tabular}}} & \multicolumn{1}{c|}{\multirow{2}{*}{\begin{tabular}[c]{@{}c@{}}\# Proxemic\\ violations\end{tabular}}} & \multicolumn{2}{c|}{Distance ratio ($\mu$ $\pm$ $\sigma$)}    & \multicolumn{2}{c|}{Time ratio ($\mu$ $\pm$ $\sigma$)}   & \multicolumn{1}{c|}{\multirow{2}{*}{\begin{tabular}[c]{@{}c@{}}Human Return \\ ($\mu$ $\pm$ $\sigma$)\end{tabular}}} \\ \cline{6-9}
\multicolumn{1}{|c|}{}                           & \multicolumn{1}{c|}{} & \multicolumn{1}{c|}{}    & \multicolumn{1}{c|}{}   & \multicolumn{1}{c|}{}       & \multicolumn{1}{c|}{Robot}         & \multicolumn{1}{c|}{Human}           & \multicolumn{1}{c|}{Robot}                         & \multicolumn{1}{c|}{Human}       & \multicolumn{1}{c|}{}         \\ \hline \hline
\multirow{3}{*}{HO}                & SA (k=0)   &    $2$  & $3$   & $177$             & \multicolumn{1}{c|}{$1.24 \pm 0.24$}             & $1.03 \pm 0.04$                   & \multicolumn{1}{c|}{${1.06} \pm {0.23}$}          & $1.10 \pm 0.13$                   & $- 1.50 \pm 2.36$                     \\ \cline{2-10}
                                            & SA (k=15)   &    $1$  & $0$   & $554$             & \multicolumn{1}{c|}{$1.11 \pm 0.06$}             & $1.03 \pm 0.06$                   & \multicolumn{1}{c|}{${1.05} \pm {0.14}$}          & $1.12 \pm 0.16$                   & $- 1.76 \pm 3.59$                     \\ \cline{2-10}
                                            & Ours  &    $\bm{0}$ &$ \bm{0}$  & $\bm{41}$         & \multicolumn{1}{c|}{$1.14 \pm 0.04$}   & $\bm{1.03} \pm \bm{0.03}$         & \multicolumn{1}{l|}{$\bm{1.00} \pm \bm{0.11}$}              &$ \bm{1.10} \pm \bm{0.13}$         & $-\bm{1.29} \pm \bm{1.24}$           \\ \hline \hline

\multirow{3}{*}{HE}              & SA (k=0)   &    $2$  & $2$ & $197$             & \multicolumn{1}{c|}{$1.25 \pm 0.27$}    & $1.03  \pm 0.03 $                 & \multicolumn{1}{c|}{${1.08}$ $\pm$ ${0.22}$}      & $1.10  \pm 0.11$                  & $- 1.63 \pm 2.06$                     \\ \cline{2-10} 
                                            & SA (k=15)   &    $2$  & $0$ & $632$             & \multicolumn{1}{c|}{$1.11 \pm 0.05$}   & $1.03  \pm 0.03 $                 & \multicolumn{1}{c|}{${1.05}$ $\pm$ ${0.10}$}      & $1.10  \pm 0.11$                  & $- 1.93 \pm 3.34$                     \\ \cline{2-10}
                                            & Ours  &    $\bm{0}$ &  $\bm{0}$ & $\bm{34}$         & \multicolumn{1}{c|}{ $1.14 \pm 0.03$}   & $\bm{1.02}  \pm \bm{0.03}$        & \multicolumn{1}{c|}{$\bm{1.01} \pm \bm{0.09}$}    & $\bm{1.10}  \pm \bm{0.10}$        & $-\bm{1.38} \pm \bm{1.04}$              \\ \hline
\end{tabular}
\end{table*}

\section{CONCLUSIONS}
In this paper, a distinction between social collision avoidance, socially aware navigation, and socially integrated navigation approaches is proposed with regard to the exhibited social behavior of the robot. 
The proposed socially integrated navigation approach is derived from sociological definitions and an understanding of human behavior in crowded environments. 
The results show that the proposed perspective change from predefined explicit social behavior to the adaption to individual human behavior leads to better robot navigation while reducing the negative impact on all agents involved. 
Consequently, robots should have the same social understanding and decision-making as humans and thus act in a socially integrated manner.
Moreover, our approach is applicable in a scalable manner through adaptive behavior.
Future work includes an evaluation in the real world.

\addtolength{\textheight}{-1cm}   

\bibliographystyle{IEEEtran}
\bibliography{indices/references}



\end{document}